\documentclass[sigconf,screen]{acmart}
\AtBeginDocument{%
  }

\setcopyright{acmlicensed}
\copyrightyear{2018}
\acmYear{2018}
\acmDOI{XXXXXXX.XXXXXXX}
\acmConference[Conference acronym 'XX]{Make sure to enter the correct
  conference title from your rights confirmation email}{June 03--05,
  2018}{Woodstock, NY}
\acmISBN{978-1-4503-XXXX-X/2018/06}

\usepackage{multirow}         
\usepackage{booktabs}         
\usepackage[most]{tcolorbox}
\usepackage{graphicx}         
\usepackage[table]{xcolor}
\usepackage{xcolor}
\definecolor{second}{rgb}{0.8, 0.9, 1.0}   
\definecolor{first}{rgb}{1.0, 0.6, 0.6}  
\usepackage{xcolor}
\definecolor{myblue}{RGB}{150,150,230}
\usepackage[table]{xcolor}
\definecolor{posgreen}{RGB}{0, 200, 0}
\definecolor{negred}{RGB}{255, 0, 0}
\begin{document}

\title[EmoTrans]{EmoTrans: A Benchmark for Understanding, Reasoning, and Predicting Emotion Transitions in Multimodal LLMs}


\author{He Hu}
\email{huhe@gml.ac.cn}
\affiliation{%
  \institution{Shenzhen University}
  \city{Shenzhen}
  \country{China}
}

\author{Tengjin Weng}
\email{wtjdsb@gamil.com}
\affiliation{%
  \institution{Shenzhen University}
  \city{Shenzhen}
  \country{China}}

\author{Zebang Cheng}
\email{zebang.cheng@gmail.com}
\affiliation{%
  \institution{Shenzhen University}
  \city{Shenzhen}
  \country{China}}

\author{Yu Wang}
\affiliation{%
  \institution{Guangdong Laboratory of Artificial Intelligence and Digital Economy (SZ)}
  \city{Shenzhen}
  \country{China}}

\author{Jiachen Luo}
\email{jiachen.luo@qmul.ac.uk}
\affiliation{%
  \institution{ Queen Mary University of London}
  \institution{Technical University of Munich}
  \city{Munich}
  \country{Germany}}

\author{Björn Schuller}
\email{schuller@tum.de}
\affiliation{%
  \institution{Imperial College London}
  \institution{Technical University of Munich}
  \city{Munich}
  \country{Germany}}

\author{Zheng Lian}
\email{lianzheng@tongji.edu.cn}
\affiliation{%
  \institution{Tongji University}
  \city{Shanghai}
    \country{China}}

\author{Laizhong Cui}
\authornotemark[1]
\email{cuilz@szu.edu.cn}
\affiliation{%
  \institution{Shenzhen University}
  \city{Shenzhen}
  \country{China}
}

\renewcommand{\shortauthors}{Trovato et al.}

\begin{abstract}
Recent multimodal large language models (MLLMs) have shown strong capabilities in perception, reasoning, and generation, and are increasingly used in applications such as social robots and human-computer interaction, where understanding human emotions is essential. However, existing benchmarks mainly formulate emotion understanding as a static recognition problem, leaving it largely unclear whether current MLLMs can understand emotion as a dynamic process that evolves, shifts between states, and unfolds across diverse social contexts. To bridge this gap, we present EmoTrans, a benchmark for evaluating emotion dynamics understanding in multimodal videos. EmoTrans contains 1,000 carefully collected and manually annotated video clips, covering 12 real-world scenarios, and further provides over 3,000 task-specific question-answer (QA) pairs for fine-grained evaluation. The benchmark introduces four tasks, namely Emotion Change Detection (ECD), Emotion State Identification (ESI), Emotion Transition Reasoning (ETR), and Next Emotion Prediction (NEP), forming a progressive evaluation framework from coarse-grained detection to deeper reasoning and prediction. We conduct a comprehensive evaluation of 18 state-of-the-art MLLMs on EmoTrans and obtain two main findings. First, although current MLLMs show relatively stronger performance on coarse-grained emotion change detection, they still struggle with fine-grained emotion dynamics modeling. Second, socially complex settings, especially multi-person scenarios, remain substantially challenging, while reasoning-oriented variants do not consistently yield clear improvements. To facilitate future research, we publicly release the benchmark, evaluation protocol, and code at \url{https://github.com/Emo-gml/EmoTrans}.
\end{abstract}

\ccsdesc[500]{Computing methodologies~Natural language processing}
\ccsdesc[500]{Computing methodologies~Computer vision}
\ccsdesc[500]{Human-centered computing~HCI design and evaluation methods}

\keywords{Multimodal Large Language Models, Emotion Transition Understanding, Emotion Benchmark, Affective Computing, Large
Multimodal Models.}
\begin{teaserfigure}
  \includegraphics[width=\textwidth]{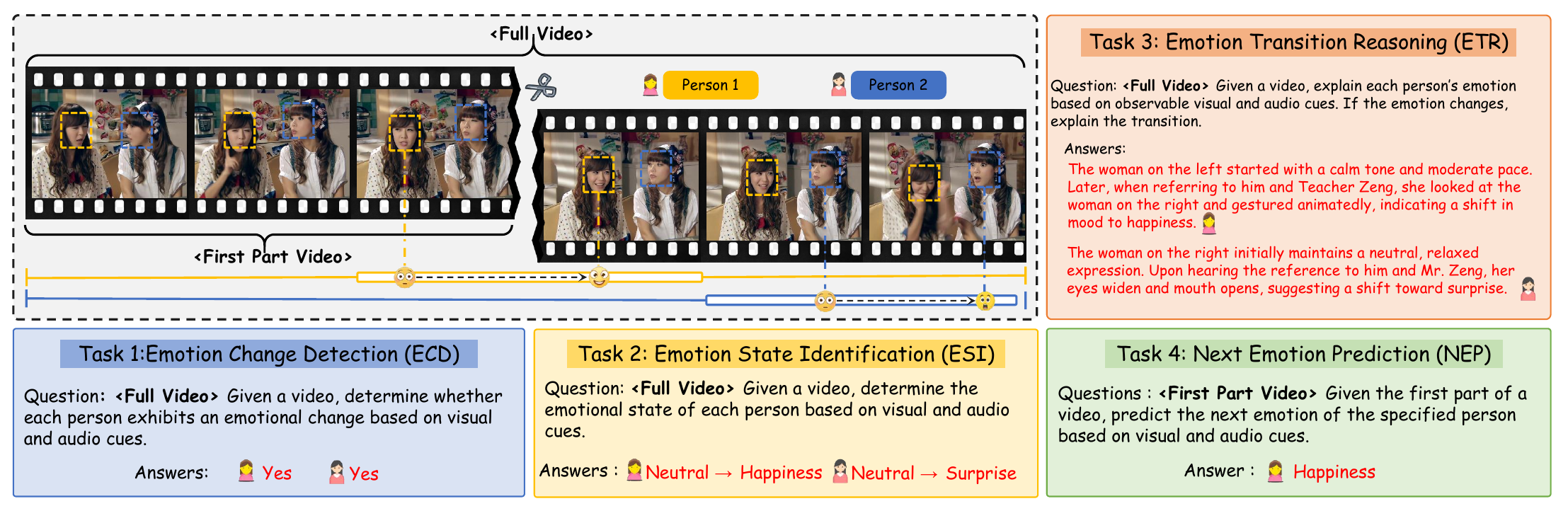}
  \caption{Overview of EmoTrans Tasks: Emotion Change Detection, Emotion State Identification, Emotion Transition Reasoning, and Next Emotion Prediction.}
  \label{fig:teaser}
\end{teaserfigure}



\maketitle

\section{Introduction}
Multimodal emotion understanding aims to recognize and interpret human emotions from diverse signals such as facial expressions, vocal tone, body movements, and contextual interactions \cite{lian2024mer, emobench, affective}. It has been extensively studied in affective computing and serves as a fundamental capability for applications including human–computer interaction, social robotics, and mental health analysis \cite{rao2025rvlf,hu2025pattern}. The field has achieved substantial progress, evolving toward more complex scenarios and emotional phenomena. Recently, multimodal large language models (MLLMs) have demonstrated remarkable capabilities in perception, reasoning, and generation, enabling a wide range of real-world applications such as embodied agents, social robots, and human–computer interaction systems \cite{yang2023emoset,emostory,fang2026emo}. As these systems become increasingly integrated into daily life, their ability to understand and respond to human emotions becomes crucial for natural and effective interaction. Although a growing body of work has explored model architectures, training strategies, and data curation to improve emotional intelligence in MLLMs, existing evaluation benchmarks are limited in their ability to assess these capabilities comprehensively. Most benchmarks formulate emotion understanding as a static prediction problem, where each instance is associated with a single dominant emotion \cite{liao2026unlocking,zhang2025videmo}. Such evaluation protocols fail to capture the inherently dynamic nature of human emotions, which evolve, transition between states, and are shaped by interactions among multiple individuals. Consequently, the extent to which current advanced MLLMs can understand, reason about, and predict emotion dynamics remains unclear under realistic and unified evaluation settings \cite{beyond,wang2026hi}.

To comprehensively assess the ability of MLLMs to understand emotion dynamics, which introduces four tasks: Emotion Change Detection (ECD), Emotion State Identification (ESI), Emotion Transition Reasoning (ETR), and Next Emotion Prediction (NEP), as illustrated in Figure~\ref{fig:teaser}. These tasks form a natural progression in difficulty, from coarse-grained detection of emotional change in ECD to increasingly fine-grained understanding, reasoning, and prediction. 
EmoTrans provides 3,274 task-specific question–answer (QA) pairs, enabling fine-grained evaluation of emotion dynamics. These QA pairs are grounded in 1,000 carefully collected and manually annotated video clips, ensuring high data quality, spanning 12 diverse real-world scenarios, and capturing rich emotion-transition patterns.
In addition, the benchmark covers both single-person and multi-person settings, supporting evaluation under diverse social interaction contexts. The main contributions are summarized as follows:
\begin{itemize}
    \item We introduce EmoTrans, a novel benchmark for evaluating emotion dynamics in multimodal videos, featuring four tasks that cover detection, identification, reasoning, and prediction of emotion transitions.
    
    \item We construct a high-quality dataset consisting of 1,000 carefully annotated video clips and 3,274 task-specific QA pairs, spanning diverse real-world scenarios and capturing rich emotion transition patterns.
    
    \item We conduct a comprehensive evaluation of 18 state-of-the-art MLLMs, revealing critical limitations in modeling temporal emotion dynamics and providing insights for future research on multimodal emotional intelligence.
\end{itemize}

\section{Related Work}
\subsection{Multimodal Emotion Understanding}
Multimodal emotion understanding has been extensively studied in affective computing, aiming to recognize human emotions from visual, acoustic, and textual signals. Early work primarily focuses on foundational emotion recognition tasks, supported by datasets such as CMU-MOSI \cite{mosi}, CMU-MOSEI \cite{mosi}, and CH-SIMS \cite{Ch-sims}, enabling progress in multimodal sentiment analysis. These studies typically formulate the task as predicting a single sentiment polarity or emotion label for each sample, focusing on static and coarse-grained emotion recognition. Beyond coarse-grained sentiment recognition, aspect-based sentiment analysis \cite{panosent} has been extended to multimodal settings to capture fine-grained sentiment toward specific aspects by jointly leveraging textual and visual signals. This line of work requires models to align multimodal cues with specific targets, enabling more precise understanding of sentiment expressions in context. Subsequent studies further investigate more complex affective phenomena, including sarcasm detection \cite{towards}, hate detection \cite{dehate}, and humor analysis \cite{hasan2019ur}, which require deeper semantic understanding and cross-modal reasoning. 

With the emergence of MLLMs, recent efforts further explore emotion reasoning and explainability. Models such as Emotion-LLaMA \cite{Emotion-llama} and AffectGPT \cite{affectgpt} attempt to generate explanations and perform affective reasoning, pushing the field from emotion recognition toward deeper emotion understanding. However, these approaches are typically formulated under simplified settings, where emotions are treated as static and tied to a single individual with a single dominant state. In contrast, real-world scenarios are inherently dynamic and often involve multiple interacting individuals, where emotions evolve, transition between states, and influence each other. Existing works lack explicit modeling of such emotion dynamics, limiting their ability to capture temporal evolution, interaction-driven changes, and future emotional trends.

\subsection{Emotion Benchmarks for MLLMs}
With the rapid advancement of MLLMs \cite{wang2025fostering}, a growing number of benchmarks have been proposed to systematically evaluate their multimodal perception and reasoning capabilities, including MMBench \cite{mmbench}, MMMU \cite{mmmu}, MME \cite{mme}, and MVBench \cite{mvbench}. These benchmarks span diverse tasks, including visual question answering, video understanding, and cross-modal reasoning, providing comprehensive assessments of general-purpose multimodal abilities across perception, cognition, and reasoning dimensions. Despite these advances, emotion-oriented evaluation remains relatively underexplored. Recent efforts such as EmoBench-M \cite{emobench}, MME-Emotion \cite{mme}, MER-UniBench \cite{affectgpt}, CA-MER \cite{benchmarking}, and MTMEUR \cite{beyond}, introduce emotion-related tasks into multimodal evaluation, extending the scope to emotion understanding and reasoning in multimodal settings. 

However, existing benchmarks primarily focus on emotion understanding through structured or progressive question answering, often emphasizing causal reasoning and response generation, while overlooking the explicit modeling of temporal emotion transitions. Moreover, most existing emotion benchmarks are constructed by reorganizing previously available datasets, which limits their ability to capture diverse and fine-grained emotional phenomena. 
To address this gap, we construct a high-quality annotated benchmark for multimodal emotion dynamics, formulating emotion change as a structured problem that encompasses detection, state identification, multimodal evidence-grounded reasoning, and prediction, thereby enabling comprehensive evaluation in realistic scenarios.


\begin{figure*}[!t]
\centering
\includegraphics[width=\linewidth]{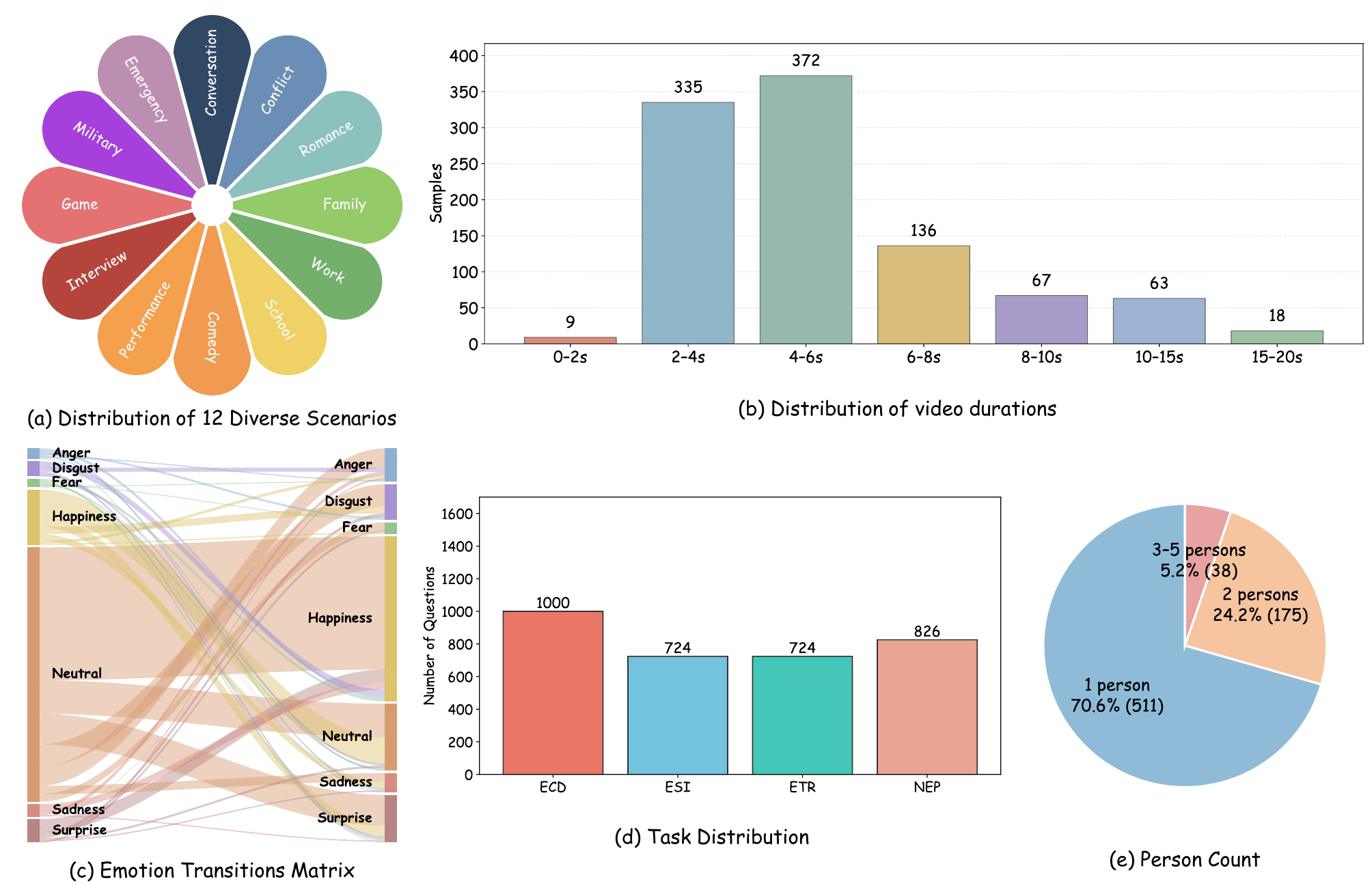}
\caption {Statistics analysis of EmoTrans.}
\label{fig:data-overview}
\end{figure*}

\section{Benchmark}
\subsection{Data Sources}
Our source data is constructed from two publicly available multimodal video corpora, OpenHumanVid \cite{openhumanvid}, and CH-SIMSv2 \cite{Ch-simsv2}. OpenHumanVid provides diverse real-world human-centric videos across various daily and social scenarios, offering rich visual and behavioral cues. CH-SIMSv2 is a widely used multimodal sentiment analysis dataset with aligned video, audio, and textual modalities; although designed for sentiment recognition, many samples exhibit multiple emotional states and temporal variations, making it suitable for studying dynamic emotion changes.

\subsection{Annotation Guidelines}

To ensure consistent and reliable annotations, we define unified guidelines covering emotion filtering, person identification, labeling, reasoning, and temporal localization.

\textbf{Emotion Change Filtering}. 
Videos are retained only if they contain clearly identifiable human subjects and sufficient multimodal cues, including facial expressions, vocal signals, and contextual interactions. 
A strict criterion is applied: at least one individual must exhibit a categorical transition between two distinct emotion classes. Variations in intensity within the same category are not considered valid changes. Samples that do not satisfy these conditions are excluded.

\textbf{Person Identification.}
Annotators identify all target individuals and assign a unique \texttt{person\_id} to each. To avoid ambiguity in multi-person scenarios, each person is described using concise and discriminative attributes, such as appearance (e.g., clothing, hairstyle), spatial position (e.g., left/right), or other salient visual features.

\textbf{Emotion Labeling}. 
Each instance is annotated based on observable multimodal cues. For videos with emotional transitions, both the initial and final states are labeled; otherwise, a single dominant emotion is assigned. All labels are drawn from seven predefined categories: Happiness, Sadness, Anger, Fear, Disgust, Surprise, and Neutral. We adopt this categorical scheme instead of dimensional representations, as discrete labels provide clearer boundaries for modeling emotion transitions and enable more reliable annotation consistency.

\textbf{Emotion Transition Reasoning}. 
Annotators provide a natural language explanation grounded in observable evidence, describing the cues that support the identified emotional state or transition. Explanations must rely only on visible and audible signals and avoid speculative inference.

\textbf{Emotion Change Localization}. 
For instance, with transitions, annotators specify the timestamp corresponding to the earliest moment at which the new emotional state becomes observable.

\subsection{Annotation Process}
The annotation process was conducted by two experts in affective
computing and three trained annotators. All annotators underwent systematic training to ensure a consistent understanding of the
guidelines. The process consists of two stages: data filtering and
annotation. Each annotated instance for a person $p_i$ is represented as:
\begin{equation}
a_i = (c_i, y_i, r_i, \tau_i),
\end{equation}
where $c_i$ denotes whether an emotion change occurs, $y_i$ denotes the annotated emotion state, $r_i$ is the reasoning, and $\tau_i$ is the timestamp of the change.

\textbf{Stage 1: Emotion Change Filtering}. 
Using approximately 15,000 candidate videos, each sample is independently assessed by three annotators. Only videos with unanimous agreement on the presence (or absence) of emotion change are retained:
\begin{equation}
c_i \in \{0,1\}.
\end{equation}
This ensures high precision in filtering. The final dataset contains 1,000 videos, including 724 with transitions and 276 without.

\textbf{Stage 2: Annotation}. 
In the second stage, annotators label emotion states, temporal change points, and reasoning explanations. The emotion annotation follows:
\begin{equation}
y_i =
\begin{cases}
(e_{\text{bef}}, e_{\text{aft}}), & c_i=1 \\
e, & c_i=0
\end{cases},
\end{equation}
where $y_i$ denotes the emotion annotation for person $p_i$, $c_i$ indicates whether an emotion change occurs, $(e_{\text{bef}}, e_{\text{aft}})$ represent the emotions before and after the change, and $e$ denotes a single stable emotion.

Experts first annotate a subset and provide detailed examples to standardize criteria. The remaining data are annotated independently in three batches, followed by expert review and feedback. Disagreements are resolved through discussion, with majority voting applied when necessary. Annotation reliability is evaluated using Fleiss’s Kappa: 0.893 for $e_{\text{bef}}$, 0.857 for $e_{\text{aft}}$, and 0.750 for paired labels, indicating substantial agreement. For temporal annotation, $\tau_i$ corresponds to the earliest observable moment of the new emotional state, with a tolerance of 0.5 seconds; larger discrepancies are resolved through expert adjudication. Experts also conduct periodic inspections to ensure consistency throughout the process.

\subsection{Dataset Statistics}

As illustrated in Figure~\ref{fig:data-overview}, EmoTrans provides a comprehensive statistical overview of the dataset across multiple dimensions. The benchmark covers 12 diverse real-world scenarios Figure~\ref{fig:data-overview}a, including conversation, conflict, interview, family, and emergency contexts, ensuring broad coverage of varied social interactions and emotional expressions. In terms of temporal distribution, Figure~\ref{fig:data-overview}b, most video clips are concentrated within the 2–6 second range, with 335 samples in 2–4 seconds and 372 samples in 4–6 seconds, while very short (0–2s) and longer clips (above 10s) are relatively limited. This distribution facilitates clearer identification of emotion transitions by focusing on concise temporal segments where changes are more localized and less ambiguous.

Regarding emotion dynamics, Figure~\ref{fig:data-overview}c, the dataset captures rich transition patterns across seven categories, with prominent transitions involving \textit{Neutral} and \textit{Happiness}, reflecting their central roles in everyday emotional expressions, while also preserving less frequent transitions such as \textit{Fear} and \textit{Disgust} to maintain diversity. EmoTrans contains a total of 3,274 task-specific QA pairs, distributed across four tasks Figure~\ref{fig:data-overview}d, including 1,000 samples for ECD, 724 for ESI, 724 for ETR, and 826 for NEP, providing sufficient coverage for both perception and reasoning capabilities. In addition, the dataset includes both single-person and multi-person settings~Figure~\ref{fig:data-overview}e, where single-person scenarios account for 70.6\%, two-person interactions for 24.2\%, and multi-person cases (3–5 individuals) for 5.2\%, enabling evaluation under both individual and interactive emotional contexts.

\begin{table*}[!t]
\caption{
Performance comparison of different methods on EmoTrans. 
A, V, and T indicate the input modalities of audio, video, and text, respectively. 
For classification tasks (ECD, ESI, and NEP), results are reported in Accuracy (\%). 
Avg. denotes the mean Accuracy over ECD, ESI, and NEP. 
\colorbox{first}{Red} and \colorbox{second}{blue} denote the best and second-best results among all models.
}
\centering
\renewcommand\arraystretch{1.2} 
\label{table_method1}
\large
\scalebox{0.95}{
\setlength{\fboxsep}{1.5pt} 
\begin{tabular}{cccccccccccc} 
\toprule
\multirow{2}{*}{Models} & \multirow{2}{*}{A} & \multirow{2}{*}{V} & \multirow{2}{*}{T} & \multirow{2}{*}{ECD} & \multirow{2}{*}{ESI} & \multicolumn{4}{c}{ETR}                      & \multirow{2}{*}{NEP}    & \multirow{2}{*} {\textbf{Avg.}} \\ \cmidrule(lr){7-10}
                        &                    &                    &                    &                      &                      & BERTScore & ROUGE-L & BLEU4  & LLM-Score &        &              \\
                         \midrule
\multicolumn{12}{c}{Open-source MLLMs} 
\\ \midrule
Qwen3-VL-8B             &                    & \checkmark                  & \checkmark                  & 71.2                & 24.4                & 87.5       & 18.4  & 3.8 & 75.4      & 27.2   & 40.9         \\
InternVL3.5-8B             &                    & \checkmark                  & \checkmark                  & 69.5                & 21.8                & 88.3       & 19.3  & 4.3  & 70.2      &20.3    & 37.2          \\
InternVL3.5-14B             &                    & \checkmark                  & \checkmark                  & 65.3                & 30.8                & 88.0       & 19.0   & 4.3 & 65.8       & 17.4    & 37.8         \\
Qwen3.5-27B              &                    & \checkmark                  & \checkmark                  & 72.8                & 38.4                & 88.3       & 19.8  & 4.4 & 77.9      & 28.1   & 46.4         \\
Qwen3-Omni-30B      & \checkmark                  & \checkmark                  & \checkmark                  & 62.9                & 23.8                & 88.9       & \colorbox{second}{\textbf{20.7}}  & \colorbox{second}{\textbf{4.6}} & 75.5      & 25.5   & 37.4         \\
Qwen3-VL-32B            &                    & \checkmark                  & \checkmark                  & 71.5                & 31.2                & 86.5       & 15.4  & 2.6 & 78.9      & 27.9   & 43.5         \\
Qwen2.5-VL-72B          &                    & \checkmark                  & \checkmark                  & 73.9                & 33.1                & 87.4       & 17.5  & 3.4 & 76.4      & 23.4   & 43.5         \\
GLM-4.5V-106B           &                    & \checkmark                  & \checkmark                  & 66.2                & 38.0                & 87.4       & 18.9  & 4.0 & 79.5      & 31.4   & 45.2         \\
GLM-4.6V-106B           &                    & \checkmark                  & \checkmark                  & 70.0                & 37.4                & 85.8       & 17.0  & 2.8 & 79.0      & 30.8   & 46.1         \\
Qwen3.5-122B       &                    & \checkmark                  & \checkmark                  & 74.0                & 38.8                & 88.4       & 20.4  & \colorbox{first}{\textbf{4.8}} & 77.3      & 28.2   & 47.0         \\
\midrule
\multicolumn{12}{c}{Closed-source MLLMs} 
\\ \midrule
Doubao-seed-2-0-lite    &                    & \checkmark                  & \checkmark                  & 64.9                & \colorbox{first}{\textbf{41.2}}                & 87.2       & 16.8  & 1.7 & 81.4      & \colorbox{first}{\textbf{36.2}} & \colorbox{second}{\textbf{47.4}} \\
Gemini-2.5-Flash        & \checkmark                  & \checkmark                  & \checkmark                  & 69.8                & 26.8                & 86.9       & 16.8  & 2.8 & \colorbox{second}{\textbf{82.7}}      & 26.2   & 40.9         \\
Gemini-3-Flash          & \checkmark                  & \checkmark                  & \checkmark                  & 73.7                & 38.3                & \colorbox{second}{\textbf{89.0}}       & 20.2  & 3.4 & 82.3      & 28.5   & 46.8         \\
Gemini-3.0-Pro            & \checkmark                  & \checkmark                  & \checkmark                  & \colorbox{second}{\textbf{74.7}}                & 36.5                & \colorbox{first}{\textbf{89.2}}       & \colorbox{first}{\textbf{21.0}}  & 4.4 & 82.3      & 28.9   & 46.7         \\
GPT-5.4                 &                    & \checkmark                  & \checkmark                  & \colorbox{first}{\textbf{75.4}}                & \colorbox{second}{\textbf{40.5}}                & 88.0       & 18.8  & 4.1 & \colorbox{first}{\textbf{85.7}}      & \colorbox{second}{\textbf{32.0}} & \colorbox{first}{\textbf{49.3}}  \\ \midrule
\end{tabular}}
\end{table*}

\subsection{Benchmark Construction}

We construct four tasks to evaluate MLLMs on multimodal emotion dynamics, covering detection, identification, reasoning, and prediction. 
Each instance is transformed into a QA pair via task-specific prompts, where the video and person information are provided as input and the expected outputs are defined accordingly. 
All inputs to MLLMs are formatted as prompts that integrate video content with structured instructions. 
Each instance may involve one or multiple persons, where $v$ denotes the input video and $\{p_1, \dots, p_n\}$ denotes the set of target persons.

\textbf{Task 1: Emotion Change Detection.}
We first consider a coarse-grained perception task that determines whether each person undergoes an emotion change:
\begin{equation}
\text{MLLM}(v, \{p_1, \dots, p_n\}) \rightarrow \{c_1, \dots, c_n\},\quad c_i \in \{0,1\}.
\end{equation}
where $c_i$ denotes whether person $p_i$ exhibits an emotion change, with $1$ indicating a change and $0$ otherwise.

\textbf{Task 2: Emotion State Identification.}
Building upon detection, we further require the model to identify the emotional state for each person:
\begin{equation}
\text{MLLM}(v, \{p_1, \dots, p_n\}) \rightarrow \{y_1, \dots, y_n\},
\end{equation}
where
\begin{equation}
y_i =
\begin{cases}
(e_{\text{bef}}, e_{\text{aft}}), & c_i=1\\
e, & c_i=0
\end{cases}.
\end{equation}

\textbf{Task 3: Emotion Transition Reasoning.}
Building upon state identification, we further require the model to generate explanations for each person:
\begin{equation}
\text{MLLM}(v, \{p_1, \dots, p_n\}) \rightarrow \{r_1, \dots, r_n\},
\end{equation}
where $r_i$ denotes the natural language rationale for person $p_i$. 
If $c_i=1$, $r_i$ explains the cause of the emotion transition from $e_{\text{bef}}$ to $e_{\text{aft}}$; 
otherwise, $r_i$ explains why the emotion $e$ remains unchanged. 
All explanations are grounded in observable multimodal evidence.
\textbf{Task 4: Next Emotion Prediction.}
Finally, we introduce a predictive setting where only the video segment before the change is provided, and the model is required to anticipate the future emotion:
\begin{equation}
\text{MLLM}(v_{<\tau}, p) \rightarrow e_{\text{next}}.
\end{equation}
Where $v_{<\tau}$ denotes the video segment before the annotated change timestamp $\tau$, $p$ denotes the target person, and $e_{\text{next}}$ denotes the predicted future emotion after $\tau$. Each instance is constructed in a person-centric manner by splitting the video according to the change timestamp.
These four tasks establish a progressive evaluation framework for multimodal emotion dynamics, moving from perception and identification to reasoning and future prediction.
\section{Experimental Settings}
\subsection{Task Formulation}
We evaluate all MLLMs in a zero-shot setting on EmoTrans to assess their ability to model emotion dynamics in multimodal videos. For
classification tasks, models are prompted to predict categories from
multimodal inputs, while for generative tasks, they are required
to provide reasoning. Depending on model capabilities, the video input is provided either as the full video or as a sequence of sampled frames, together with the associated text.

\subsection{Evaluation Models and Metrics}
We evaluate the performance on a total of 18 cutting-edge MLLMs, including GLM-4.5V/4.6V~\cite{chatglm}, InternVL3.5-8B/14B~\cite{internvl3}, Qwen2.5-VL-72B~\cite{25vltechnicalreport}, Qwen3-VL-8B/32B~\cite{qwen3}, Qwen3.5-27B/122B~\cite{qwen3_5}, Qwen3-Omni-30B~\cite{Qwen3-omni}, Doubao-seed-2.0-lite~\cite{doubao_seed_2_0_lite}, Gemini-2.5-Flash~\cite{team2024gemini}, Gemini-3.0-Flash~\cite{team2024gemini}, Gemini-3.0-Pro~\cite{team2024gemini}, and GPT-5.4~\cite{singh2025openai}. We evaluate classification tasks using Accuracy, while for generation tasks we adopt a combination of lexical and semantic metrics, including BLEU-4 \cite{zhang2019bertscore}, ROUGE-L \cite{papineni2002bleu}, and BERTScore \cite{lin2004rouge}; additionally, we incorporate an LLM-based evaluation protocol by employing a strong text-only large language model GPT-5 \cite{singh2025openai} as a judge to assess whether the generated explanation semantically matches the ground-truth explanation, focusing on consistency in underlying emotional reasoning rather than surface-level similarity.

\begin{table}[!t]
\centering
\renewcommand\arraystretch{1.25}
\caption{Comparison of standard and thinking models in terms of the average Accuracy (\%) across ECD, ESI, and NEP, where the Avg. is computed as the mean Accuracy over the three tasks.}
\label{tab:thinking_avg}
\large
\begin{tabular}{lc}
\toprule
\textbf{Model} & \textbf{Avg.} \\
\midrule
Qwen3-VL-32B & 43.53 \\
Qwen3-VL-32B-Thinking & 43.57 (\textcolor{posgreen}{\scriptsize$\uparrow +0.04$}) \\
\midrule
Gemini-2.5-Flash & 40.93 \\
Gemini-2.5-Flash-Thinking & 42.92 (\textcolor{posgreen}{\scriptsize$\uparrow +1.99$}) \\
\midrule
Gemini-3.0-Flash & 46.83 \\
Gemini-3.0-Flash-Thinking & 47.00 (\textcolor{posgreen}{\scriptsize$\uparrow +0.17$}) \\
\bottomrule
\end{tabular}
\end{table}

\begin{figure}[!t]
\centering
\includegraphics[width=\linewidth]{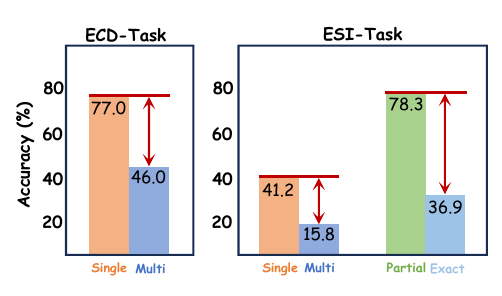}
\vspace{-7mm}
\caption{Performance comparison under different evaluation settings. 
\textbf{Single} and \textbf{Multi} denote single-person and multi-person scenarios, respectively. 
In the ESI task, \textbf{Partial} evaluates whether at least one emotion state is correctly identified, 
while \textbf{Exact} requires both emotion states to be correctly predicted.}
\label{fig:Comparison}
\end{figure}

\vspace{-5mm}

\vspace{3mm}
\subsection{Main Results}
Table ~\ref{table_method1} summarizes the performance of various MLLMs on the EmoTrans task. Overall, closed-source models consistently outperform open-source models on most tasks. Among open-source models, Qwen3.5-122B achieves the best overall performance with an average score of 47.0\%, showing relatively stronger capability in modeling emotion transitions. Among proprietary models, GPT-5.4 achieves the highest average score of 49.3\%. Despite this, overall performance remains limited, with no model exceeding 50\% average accuracy. In particular, most models score below 40\% on ESI, indicating the difficulty of fine-grained emotion state identification. From a modality perspective, several MLLMs using only visual and textual inputs achieve competitive performance, while fully multimodal MLLMs that incorporate audio, visual, and textual information tend to show lower performance. From a task perspective, among the 18 evaluated models, the results show clear differences in task difficulty across ECD, ESI, and NEP, with performance decreasing progressively from ECD to NEP. In particular, for the more challenging NEP task, even the best-performing MLLM achieves only around 36\% accuracy. Overall, these results highlight the challenging nature of EmoTrans and indicate that current MLLMs remain limited in modeling dynamic emotion.

We further analyze model performance under different settings. As shown in Figure~\ref{fig:Comparison}, multi-person scenarios are significantly more challenging than single-person cases, with performance dropping from 77.0\% to 46.0\% on ECD and from 41.2\% to 15.8\% on ESI, indicating that interpersonal interactions introduce substantial ambiguity. For ESI, models achieve much higher performance under partial matching, while performance drops significantly when requiring the complete transition to be correct (74.0\% vs. 35.6\%). This suggests that models struggle to capture full emotion transitions, reflecting limitations in modeling temporal dynamics and fine-grained emotional changes. Furthermore, as shown in Table~\ref{tab:thinking_avg}, thinking variants bring only marginal improvements, indicating that enhanced reasoning does not consistently translate into better performance on emotion dynamics tasks.

\subsection{Observations and Insights}
From the experimental results, we draw the following key observations and insights.

\textbf{Significant Performance Gap Between Detection and Deeper Emotion Understanding.}
The results reveal a clear gap between coarse-grained detection and deeper emotion-dynamics understanding. While ECD is comparatively easier, performance drops notably on ESI, ETR, and NEP, which require models to identify specific emotional states, explain transitions, or anticipate future changes. This suggests that current MLLMs can often perceive that an emotional shift has occurred, but still struggle to characterize its content, cause, and temporal development accurately.

\textbf{Fully Multimodal Models Do Not Necessarily Achieve Superior Performance.}
Fully multimodal models do not consistently achieve the best results across all tasks. Although access to visual, acoustic, and textual information provides richer evidence, the benefit is not automatic. This finding suggests that the core challenge lies less in modality coverage itself and more in effective cross-modal alignment, selective cue integration, and temporally grounded emotion understanding.

\textbf{Multi-Person Scenarios Remain Significantly More Challenging.}
Multi-person settings are consistently more difficult than single-person ones. In such cases, models must not only capture emotional cues, but also correctly associate them with the corresponding individual over time. The performance gap indicates that person-specific temporal grounding and cross-person disambiguation remain major bottlenecks for current MLLMs in socially interactive scenarios.

\textbf{Thinking Models Provide Limited and Inconsistent Benefits.}
Thinking variants bring only modest and inconsistent improvements. Although they may slightly improve average performance on some tasks, such gains are generally small and do not consistently translate into stronger emotion understanding. This suggests that the main bottleneck of emotion dynamics modeling may lie more in multimodal perception and temporal grounding than in generic reasoning depth alone.

\subsection{Conclusion}
In this paper, we present EmoTrans, a benchmark for evaluating emotion dynamics understanding in MLLMs. EmoTrans moves beyond static emotion recognition and introduces four tasks to assess emotion change detection, state identification, transition reasoning, and future emotion prediction in multimodal videos. Experiments on 18 state-of-the-art MLLMs show that current models can handle coarse-grained emotion change detection to some extent, but still struggle with fine-grained emotion dynamics modeling, especially in socially complex settings such as multi-person scenarios. These results highlight the gap between current MLLMs and robust emotion dynamics understanding. In the future, we plan to expand the benchmark to broader scenarios and richer interaction settings, and further explore more reliable evaluation protocols for dynamic emotion understanding.


\bibliographystyle{ACM-Reference-Format}
\bibliography{samples/sample-base}

\clearpage
\appendix

\section{Evaluation Settings Analysis for ECD and ESI}\label{app:further_analyze}
Table~\ref{table_method} presents the performance of both open-source and closed-source MLLMs on the EmoTrans benchmark across ECD and ESI tasks. Overall, closed-source models demonstrate stronger performance than open-source models in most settings, particularly in more challenging multi-person scenarios. Across tasks, a clear performance gap can be observed between single-person and multi-person settings, indicating that modeling emotion dynamics in multi-person interactions remains a significant challenge. For ESI, the large discrepancy between Partial Match and Exact Match further highlights the difficulty of accurately capturing both source and target emotional states. Although some open-source models achieve competitive results on certain metrics, their performance is generally less stable compared to closed-source models. In addition, no model achieves consistently high performance across all settings, suggesting that emotion dynamics understanding remains an open and challenging problem for current MLLMs.

\section{Thinking Variant Analysis}
To further investigate the impact of enhanced reasoning mechanisms, we compare standard MLLMs with their corresponding thinking variants on the EmoTrans benchmark. Table~\ref{table_thinking} reports the average performance across four tasks, including ECD, ESI, ETR, and NEP. Thinking variants bring limited and inconsistent improvements across tasks. While slight gains can be observed in certain cases, such as ESI or NEP for specific models, these improvements are neither stable nor generalizable. In some cases, thinking variants even lead to performance degradation, particularly on ETR, suggesting that increased reasoning depth does not necessarily translate into better emotion understanding. These results indicate that the primary bottleneck in emotion dynamics modeling may not lie in high-level reasoning, but rather in multimodal perception and temporal grounding. Simply scaling reasoning processes is insufficient to address the challenges of fine-grained emotion transition understanding.

\section{Task Examples}
\label{sec:task_examples}
\begin{figure*}[t]
\centering
\includegraphics[width=\linewidth]{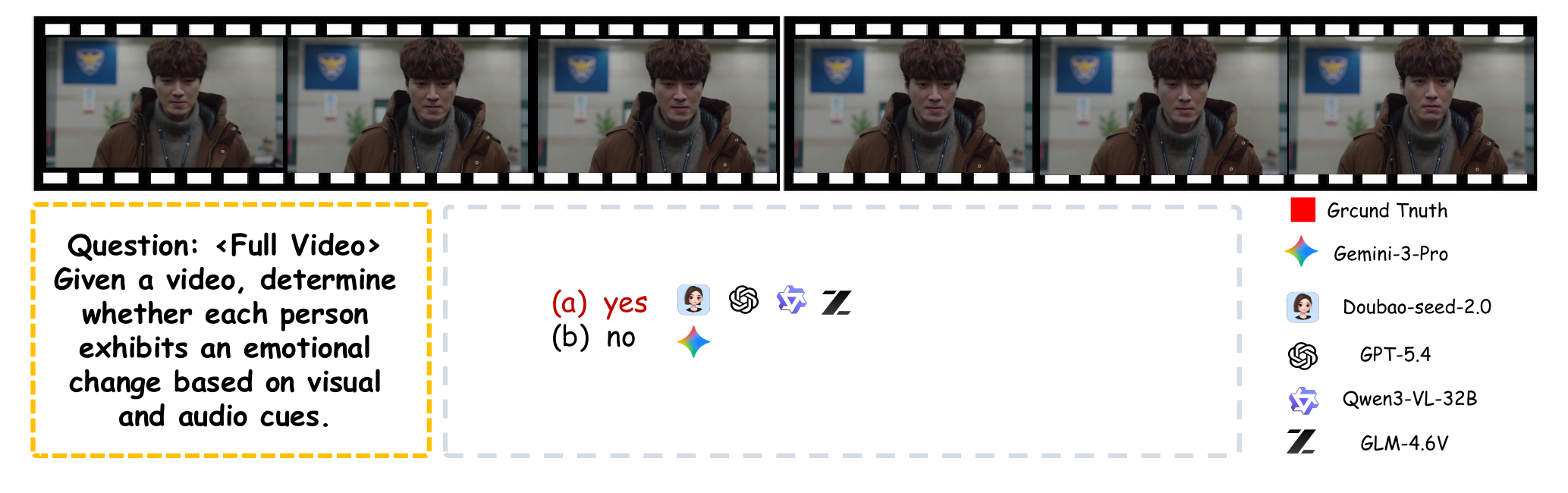}
\caption{
\textbf{Example of Emotion Change Detection (ECD).} 
}
\label{fig:ecd}
\end{figure*}

\begin{figure*}[t]
\centering
\includegraphics[width=\linewidth]{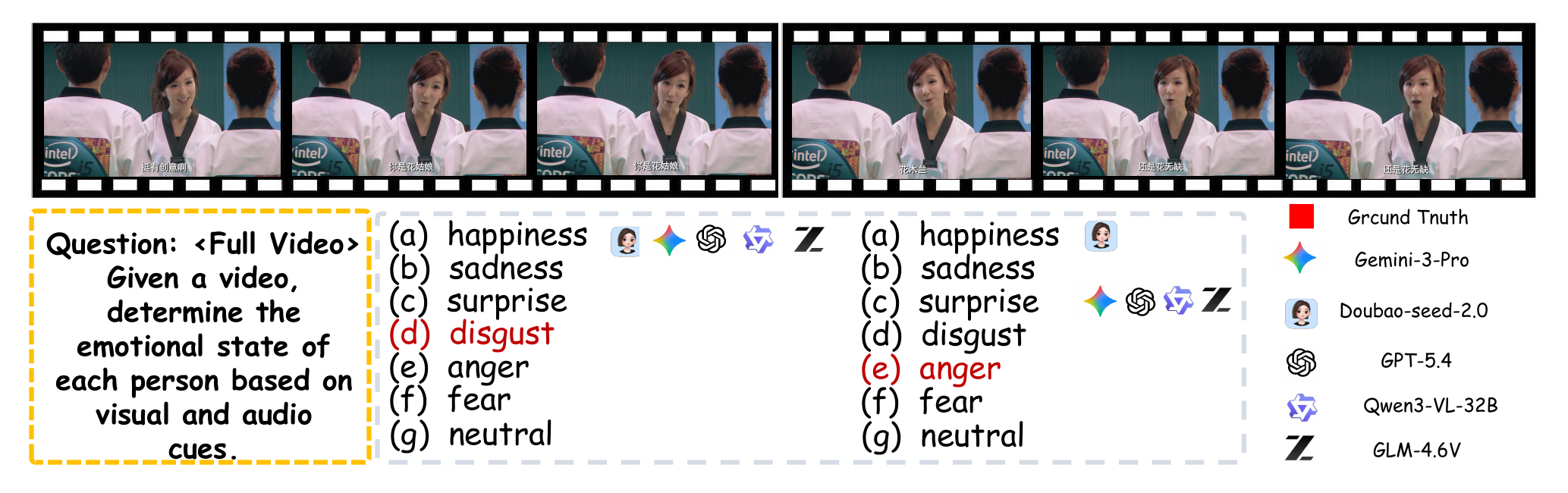}
\caption{
\textbf{Example of Emotion State Identification (ESI).} 
}
\label{fig:esi}
\end{figure*}

\begin{figure*}[t]
\centering
\includegraphics[width=\linewidth]{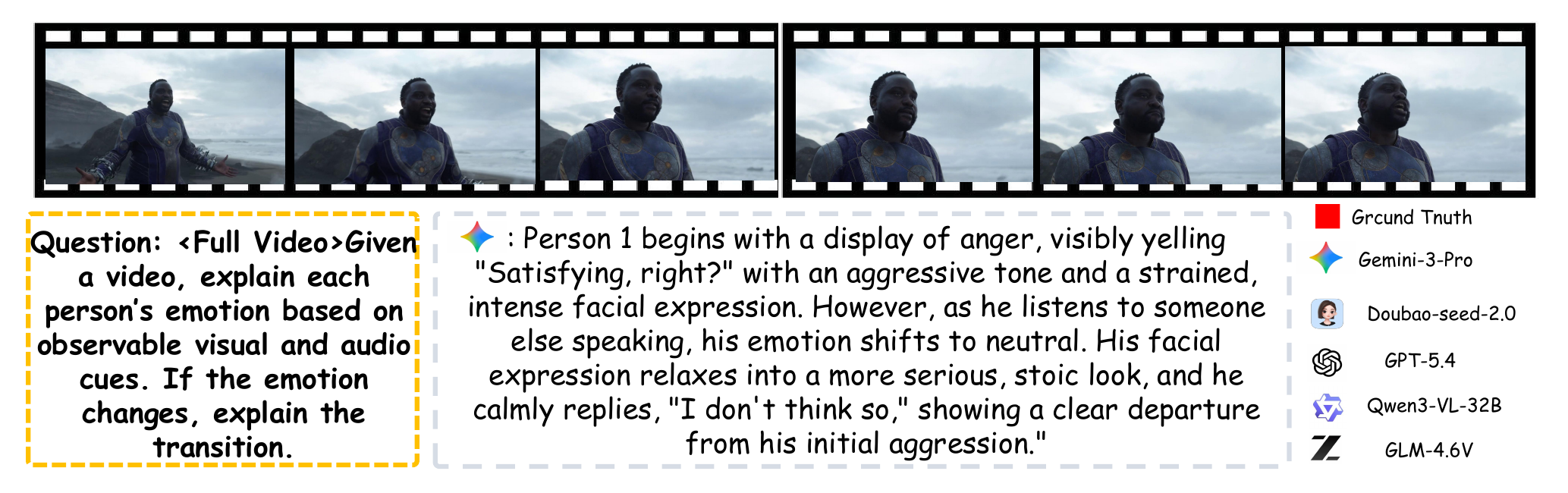}
\caption{
\textbf{Example of Emotion Transition Reasoning (ETR).} 
}
\label{fig:etr}
\end{figure*}

\begin{figure*}[t]
\centering
\includegraphics[width=\linewidth]{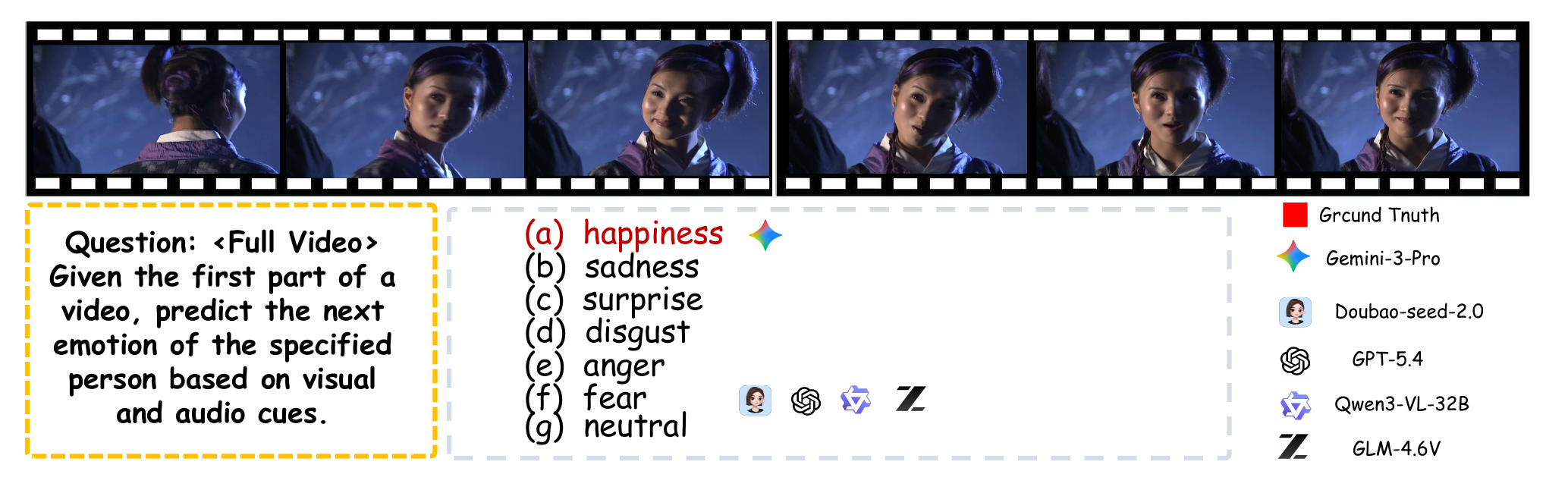}
\caption{
\textbf{Example of Next Emotion Prediction (NEP).} 
}
\label{fig:nep}
\end{figure*}

Illustrative examples of the four tasks are presented in Figures~\ref{fig:ecd}--\ref{fig:nep}, corresponding to Emotion Change Detection (ECD), Emotion State Identification (ESI), Emotion Transition Reasoning (ETR), and Next Emotion Prediction (NEP), respectively.

\begin{table*}[!t]
\centering
\renewcommand\arraystretch{1.2} 
\caption{
Performance comparison of multimodal large language models on emotion dynamics tasks.
ECD and ESI denote Emotion Change Detection and Emotion State Identification, respectively.
Single and Multi refer to single-person and multi-person scenarios.
For ESI, Partial Match measures whether at least one emotion state is correctly identified,
while Exact Match requires both source and target emotions to be correctly predicted.
All results are reported in Accuracy (\%).
Best and second-best results are highlighted in red and blue, respectively.
}
\label{table_method}
\large
\scalebox{0.9}{
\setlength{\fboxsep}{1.5pt} 
\begin{tabular}{ccccccc} 
\midrule
\multirow{2}{*}{Models} & \multicolumn{2}{c}{ECD} & \multicolumn{4}{c}{ESI} \\ \cmidrule(lr){2-3} \cmidrule(lr){4-7}
                        & Single  & Multi & Single & Multi & Partial Match & Exact Match \\ \midrule
\multicolumn{7}{c}{Open-source MLLMs} \\ \midrule
Qwen3-VL-8B             & 78.5  & 44.6  & 33.7 & 2.3   & 66.2          & 28.2 \\
Qwen3.5-27B             & 77.6  & 55.7  & 47.8 & 17.5  & 81.8          & 43.1 \\
Qwen3-Omni-30B-A3B      & 69.5  & 38.5  & 30.7 & 7.0   & 69.2          & 26.5 \\ 
Qwen3-VL-32B            & 77.4  & 49.8  & 38.5 & 13.6  & 79.0          & 36.6 \\
Qwen2.5-VL-72B          & \colorbox{second}{\textbf{81.2}} & 46.9  & 40.3 & 16.0  & 79.1          & 37.2 \\
GLM-4.5V-106B            & 72.7  & 42.3  & 45.8 & 19.2  & 82.6          & 41.6 \\
GLM-4.6V-106B            & 76.8  & 44.1  & 45.0 & 19.2  & 82.5          & 40.6 \\
Qwen3.5-122B-A10B       & 78.7  & \colorbox{first}{\textbf{56.8}}  & 46.4 &20.7  & \colorbox{first}{\textbf{83.8}} & 43.2 \\
 \midrule
 \multicolumn{7}{c}{Closed-source MLLMs} \\ \midrule
Doubao-seed-2-0-lite    & 69.3  & 48.8  & \colorbox{first}{\textbf{48.9}} & \colorbox{first}{\textbf{22.5}}  & 81.8          & \colorbox{second}{\textbf{45.4}} \\
Gemini-2.5-Flash        & 78.0  & 39.4  & 34.4 & 8.4   & 80.7          & 28.5 \\
Gemini-3-Flash          & \colorbox{first}{\textbf{82.0}}  & 43.2  & 46.0 & 19.8  & 83.4           & 41.4  \\
Gemini-3-Pro            & \colorbox{first}{\textbf{82.0}}  & 47.9  & 43.4 & 19.8  & \colorbox{second}{\textbf{83.5}} & 39.6 \\
GPT-5.4                 & 80.6  & \colorbox{second}{\textbf{56.3}}  & \colorbox{second}{\textbf{48.3}} &  \colorbox{second}{\textbf{21.6}}  & 82.7          & \colorbox{first}{\textbf{46.8}} \\ \midrule
\end{tabular}}
\end{table*}

\begin{table*}[!t]
\centering
\renewcommand\arraystretch{1.2} 
\caption{
Effect of thinking variants on EmoTrans across four tasks.
ECD, ESI, ETR, and NEP denote Emotion Change Detection, Emotion State Identification, Emotion Transition Reasoning, and Next Emotion Prediction, respectively.
All results are reported in Accuracy (\%).
“Thinking” models refer to variants with enhanced reasoning processes.
The reported scores are averaged over all evaluation settings for each task.
}
\label{table_thinking}
\large
\scalebox{0.9}{
\setlength{\fboxsep}{1.5pt} 
\begin{tabular}{lcccc}
\toprule
\textbf{Model} & \textbf{ECD} & \textbf{ESI} & \textbf{ETR} & \textbf{NEP} \\ 
\midrule
Qwen3-VL-32B & 71.50 & 31.20 & 78.90 & 27.90 \\
Qwen3-VL-32B-thinking & 70.50 & 33.43 & 81.70 & 26.79 \\ \midrule
\addlinespace
Gemini-2.5-Flash & 69.80 & 26.80 & 82.70 & 26.20 \\
Gemini-2.5-Flash-thinking & 70.11 & 28.71 & 75.50 & 29.94 \\ \midrule
\addlinespace
Gemini-3-Flash & 73.70 & 38.30 & 82.30 & 28.50 \\
Gemini-3-Flash-thinking & 70.46 & 38.70 & 64.70 & 31.84 \\
\bottomrule
\end{tabular}}
\end{table*}

\section{Prompt}\label{app:prompt}

\textbf{Task 1: Emotion Change Detection (ECD)}
\begin{tcolorbox}[
    colback=myblue!5!white,
    colframe=myblue!75!black,
    arc=1mm, 
    auto outer arc,
    title={Prompt for Emotion Change Detection},
    breakable
]\small

Given a video and a list of persons, determine whether each person exhibits an emotion transition at any point in the video.

Base your judgment only on observable visual and audio cues (e.g., facial expressions, tone of voice, body language, actions, and subtitles if provided). Analyze each person independently.

Do not provide any explanation or additional content.



\end{tcolorbox}

\vspace{5em}

\textbf{Task 2: Emotion State Identification (ESI)}
\begin{tcolorbox}[
    colback=myblue!5!white,
    colframe=myblue!75!black,
    arc=1mm, 
    auto outer arc,
    title={Prompt for Emotion State Identification},
    breakable
]\small

Given a video and a list of persons, determine the emotional transition of each person.

Use only observable visual and audio cues (facial expressions, tone of voice, body language, actions, and subtitles if provided). Analyze each person independently.

Allowed emotion labels:  
Happiness, Sadness, Anger, Fear, Disgust, Surprise, Neutral

If no change occurs, the starting and ending emotions must be identical.

Do not provide any explanation or additional content.



\end{tcolorbox}

\vspace{0.5em}

\textbf{Task 3: Emotion Transition Reasoning (ETR)}
\begin{tcolorbox}[
    colback=myblue!5!white,
    colframe=myblue!75!black,
    arc=1mm, 
    auto outer arc,
    title={Prompt for Emotion Transition Reasoning},
    breakable
]\small

Given a video and a list of persons, explain the emotion of each person.

If an emotion transition occurs, explain why the emotion shifts from the starting state to the ending state. Otherwise, explain why the emotion remains unchanged.

Use only observable visual and audio cues (facial expressions, tone of voice, body language, actions, and subtitles if provided). Do not introduce unobservable information.


\end{tcolorbox}

\vspace{0.5em}

\textbf{Task 4: Next Emotion Prediction (NEP)}
\begin{tcolorbox}[
    colback=myblue!5!white,
    colframe=myblue!75!black,
    arc=1mm, 
    auto outer arc,
    title={Prompt for Next Emotion Prediction},
    breakable
]\small

Given a partial video and a specified person, predict the next emotion.

Use only observable visual and audio cues (facial expressions, tone of voice, body language, actions, and subtitles if provided). Analyze only the given person.

Allowed emotion labels:  
Happiness, Sadness, Anger, Fear, Disgust, Surprise, Neutral

Do not provide any explanation or additional content.


\end{tcolorbox}

\end{document}